\documentclass[12pt]{article}

\usepackage{amsmath}
\usepackage{tikz-cd}
\usepackage{amssymb}
\usepackage{graphicx}
\usepackage{hyperref}
\usepackage{geometry}
\usepackage{mathtools}
\usepackage{titlesec}
\usepackage{titling}
\usepackage[backend=bibtex,style=numeric,sorting=none,defernumbers=true]{biblatex}
\usepackage{etoolbox}   

\titleformat{\section}      {\normalfont\Large\bfseries\centering}{\thesection}{1em}{}
\titleformat{\subsection}   {\normalfont\large\bfseries\centering}{\thesubsection}{1em}{}
\titleformat{\subsubsection}{\normalfont\normalsize\bfseries\centering}{\thesubsubsection}{1em}{}

\titleclass{\subsubsubsection}{straight}[\subsubsection]
\newcounter{subsubsubsection}[subsubsection]
\renewcommand\thesubsubsubsection{\thesubsubsection.\arabic{subsubsubsection}}
\titleformat{\subsubsubsection}
  {\normalfont\normalsize\bfseries\centering}{\thesubsubsubsection}{1em}{}
\titlespacing*{\subsubsubsection}{0pt}{1.5ex plus 0.5ex minus .2ex}{1em}

\pretitle{%
  \begin{center}
    \vspace*{-0.42cm}
    \rule{\linewidth}{3pt}\\[\smallskipamount]
    \bfseries\LARGE
    \vspace{0.22cm}
}
\posttitle{%
    \\[\smallskipamount]
    \rule{\linewidth}{1.6pt}
  \end{center}
  \vspace{0.12cm}
}

\preauthor{\begin{center}\large}
\postauthor{\end{center}}

\predate{\begin{center}\large}
\postdate{\end{center}}

\makeatletter
\g@addto@macro\maketitle{\thispagestyle{empty}}
\makeatother

\title{Gauge Flow Models}

\author{%
  \begin{minipage}[t]{0.46\textwidth}\centering
    Alexander Strunk\thanks{Corresponding author: \href{mailto:astrunk.research@evercot.ai}{astrunk.research@evercot.ai}}\\
    Evercot AI\\
  \end{minipage}%
  \hspace{1em}
  \begin{minipage}[t]{0.46\textwidth}\centering
    Roland Assam\\
    Evercot AI\\
  \end{minipage}%
\vspace{0.32cm}
}

\date{19 June 2025}

\begin{document}

\maketitle  

\vspace{-0.80cm} 
\begin{abstract}
\noindent
This paper introduces Gauge Flow Models, a novel class of 
Generative Flow Models.  These models incorporate a learnable Gauge Field within the Flow Ordinary Differential Equation (ODE).  A comprehensive 
mathematical framework for these models, detailing their construction and properties, is provided.  Experiments using Flow Matching on Gaussian Mixture Models demonstrate that Gauge Flow Models yield significantly better performance than traditional Flow Models of comparable or even larger size.  Additionally, unpublished research indicates a potential for enhanced  performance across a broader range of generative tasks.
\end{abstract}

\newpage
\setcounter{page}{1}

\section{Introduction}
A Gauge Flow Model is defined within the mathematical framework of associated bundles:
\begin{align*}
\hat{A} = P \times_{G} F
\end{align*} 
where $P = (P, M, G, \pi_P)$ is a principal bundle with structure group $G$ and $F$ is a $G$-space.
The dynamics are governed by the following neural ODE:
\begin{align*}
\hat{\nabla}_{dt} x(t):= v_{\theta}(x(t),  t) - \alpha(t) \Pi_{M} ( A_{\mu
}(x(t),  t) d^{\mu}(x(t), t)  v_{s}(x(t), t) )
\end{align*} 
This equation incorporates several novel key components: 
\begin{itemize}
\item $v_{\theta}(x(t),  t) \in TM$ is a learnable vector field,  defined as a section of the tangent bundle over the base manifold $M$.
\item $\alpha(t)$ represents a learnable schedule.
\item $A_{\mu}(x(t),  t)$ is a learnable Gauge Field valued in the Lie algebra of the Gauge Group $G$.
\item $d^{\mu}(x(t), t)$ is a direction vector field,  modeled as a smooth section of the tangent bundle $TM$.
\item $v_{s}(x(t), t)$ is a further learnable vector field corresponding to a section on the associated bundle $P \times_{G} F$.
\item $\Pi_{M}: P \times_{G} F \to TM$ is a smooth,  learnable projection map. 
\end{itemize}
The novelty of this Flow ODE lies in the inclusion of the Gauge Term, 
which introduces a novel combination of learnable and non-learnable vector 
fields – a feature absent from standard Neural Flow Models.  The standard 
Neural Flow Model,  as defined by \cite{CNFs, FM},  is characterized by:
\begin{align*}
\frac{dx(t)}{dt} = v_{\theta}(x(t),  t) 
\end{align*} 
\noindent The Gauge Field associated with a predefined Gauge Group (e.g., $SO(N)$ or $SU(N)$) establishes a crucial geometric inductive bias.  By explicitly incorporating this bias,  the model learns to represent 
data more efficiently,  favoring arrangements that align with the imposed 
symmetries.  Consequently,  these models often achieve 
stronger performance and improved robustness compared to models lacking 
this geometric constraint. This approach is particularly advantageous in domains like protein or 
drug design, where molecules frequently exhibit rotational or 
translational symmetries.
 \newpage

\section{Mathematical Background}
This section introduces the mathematical foundations of Gauge Flow 
Models, primarily utilizing the concept of Fiber Bundles. First,  the necessary mathematical tools are established, culminating in a comprehensive 
description of Gauge Flow Models. A working familiarity with concepts from 
Differential Geometry – including manifolds, vector fields, differential 
forms, and Lie groups – is assumed. For readers seeking a refresher or 
introduction to these topics, several relevant resources are listed below.
\begin{itemize}
\item \textbf{Differential Geometry}: \cite{DifferentialGeometry,  DifferentialGeometryManifoldConnections,  DifferentialGeometryManifold,  SmoothManifolds, Manifolds}
\item \textbf{Gauge Theory}: \cite{MathematicalGaugeTheory,  GeometryTopologyPhysics, FibreBundle}
\item  \textbf{Lie Groups and Lie Algebras}: \cite{LieGroupsLieAlgebras, LieAlgebras}
\end{itemize}

\subsection{Fiber Bundle}
A fiber bundle is a geometric construction that assembles a continuous family of fibers,  one for each point of a base space, in a way that varies continuously across the base.  Locally,  it resembles a 
product of the base space and the fiber.  However, it's important to note 
that while they are locally trivial, fiber bundles can generally deviate from being a 
global product space in non-trivial ways. An illustration of a fiber bundle is provided below.

\begin{figure}[ht]
  \centering
  \includegraphics[width=\textwidth]{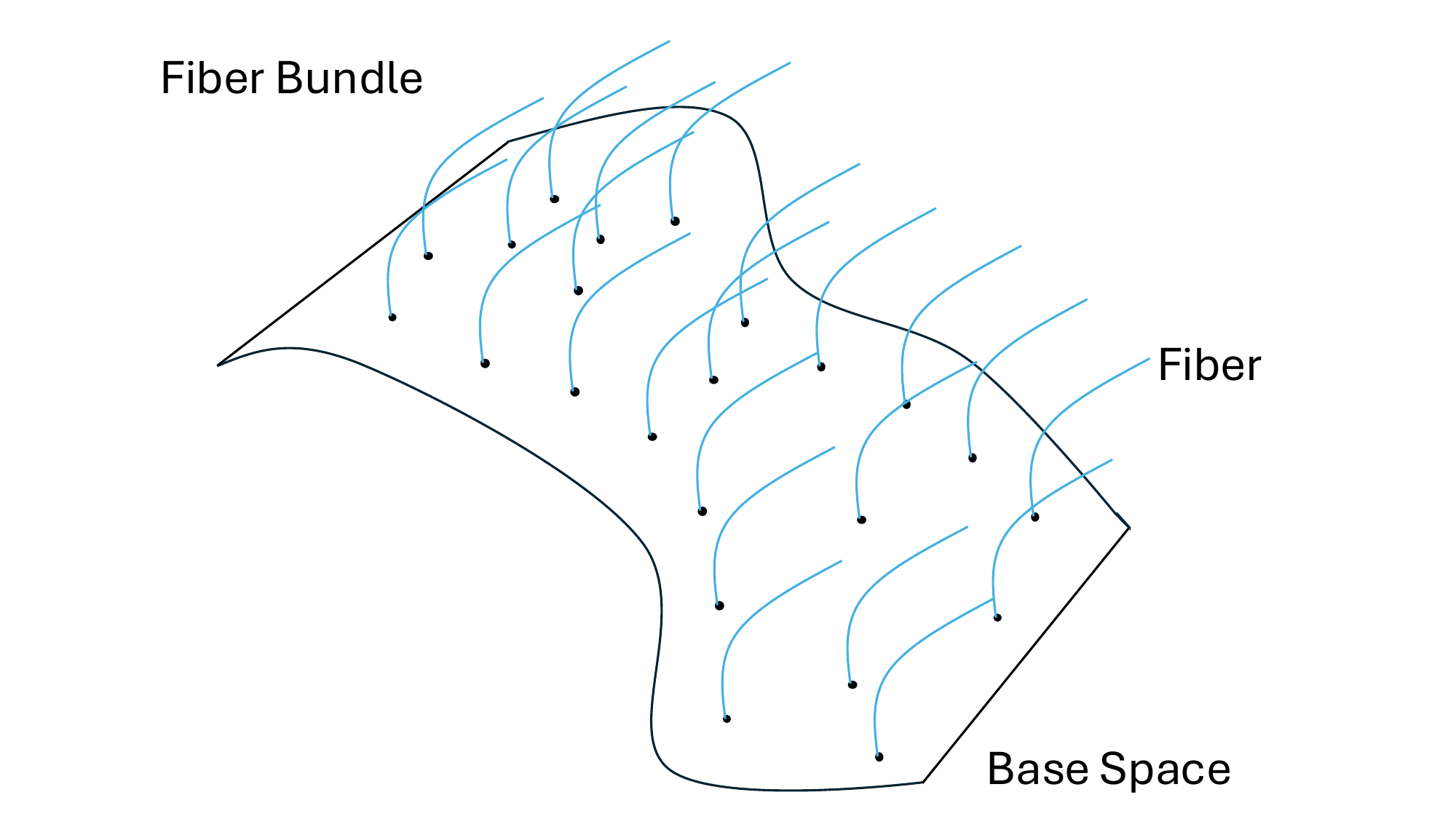}
  \caption{A fiber bundle can be visualized as a base space with a fiber attached to each point.  As shown,  the base space corresponds to the plane and the blue lines represent the fibers.}
  \label{fig:Fiber_Bundle}
\end{figure}
\newpage
\noindent A fiber bundle is formally defined as the data $(E, B,F, \pi)$,  where:
\begin{itemize}
\item \textbf{Total space}: E is the union of all the fibers.
\item \textbf{Base space}: B is the parameter space.
\item \textbf{Fiber}: F is the typical space sitting over each point of B.
\item \textbf{Projection}: $\pi: E \mapsto B$ is a continuous surjective map. 
\end{itemize}
Ensuring local triviality – meaning that the bundle behaves like a 
product locally – requires that for every point $x \in B$ there exists 
an open neighborhood $U_x \subset B$ and a fiber preserving homeomorphism $\Psi$:
\begin{align*}
\Psi: \pi^{-1}(U_x) \mapsto U_x  \times F
\end{align*}
such that the following diagram commutes:
\[
\begin{tikzcd}
\pi^{-1}(U_x) \arrow[r, "\Psi"] \arrow[d, "\pi"'] 
  & U_{x} \times F \arrow[d, "\mathrm{pr}_1"] \\
U_x \arrow[r, "\mathrm{id}_{U_x}"'] 
  & U_x
\end{tikzcd}
\]
This local triviality condition ensures that the bundle can be locally 
described as a product of the base space and the fiber.

\subsubsection{Connection on a Fiber Bundle}
A connection on a fiber bundle defines a rule for parallel transport, enabling the comparison of fibers at infinitesimally nearby points. This provides a way to transport elements along curves in the base manifold in a geometrically consistent manner.

\begin{figure}[ht]
  \centering
  \includegraphics[width=\textwidth]{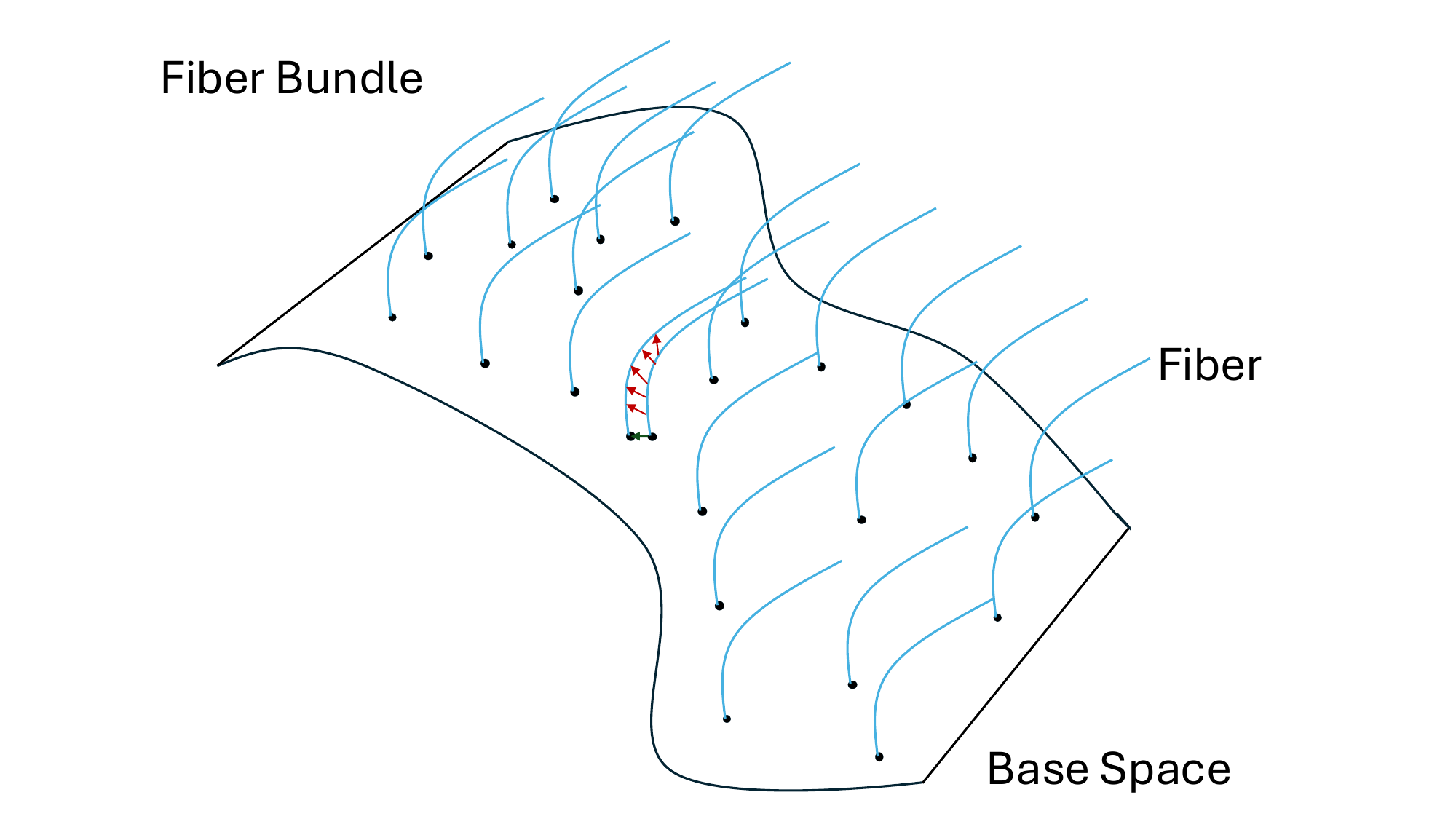}
  \caption{A connection on a fiber bundle defines a mapping that relates each fiber 
to its infinitesimally near neighboring fibers. This enables local ‘parallel transport’ within the bundle.}
  \label{fig:Fiber_Bundle_Connection}
\end{figure}
\noindent Mathematically, a connection on a fiber bundle $(E, B, F, \pi)$ is defined 
as a smoothly varying family of horizontal subspaces $H_p E$:
\begin{align*}
HE \subset TE
\end{align*}
where:
\begin{itemize}
\item[1.)] At every point $p \in E$, the tangent space $T_p E$ is split into a 
horizontal component $H_p E$ and a vertical component $V_p E$:
\begin{align*}
T_{p} E = H_{p} E \oplus V_{p}E
\end{align*}
The vertical component,  $V_{p}E$ is defined via the differential $d\pi$:
\begin{align*}
V_{p}E := ker(d\pi_p : T_{p}E \mapsto T_{\pi(p)}B)
\end{align*}
\item[2.)] The differential of the projection map $d\pi_p$, restricted to the horizontal subspace $H_p E$  is a linear isomorphism:
\begin{align*}
d\pi_p|_{H_p E}: H_p E \rightarrow T_{\pi (p) }B
\end{align*}
\end{itemize}
In essence, a connection is precisely the choice of such a smoothly 
varying family of horizontal subspaces $H_p E \subset T_p E$. This 
particular type of connection is often referred to as an Ehresmann 
Connection.

\subsubsection{Principal Bundle}
A principal bundle $P := (P, B, G, \pi)$ is a special type of fiber bundle in which the fiber is a Lie group G acting freely and transitively on itself.  Principal bundles form the geometric foundation for gauge 
theories in physics, connections in differential geometry, and much of 
modern topology.
\subsubsubsection{Lie Groups}
A Lie Group $(G, \bullet)$ is a differentiable manifold $G$ equipped with a binary operation $\bullet$:
 \begin{align*}
\bullet: &G \times G \rightarrow G \\
&(g,h) \mapsto g \bullet h  
\end{align*}
such that the following group axioms are satisfied: 
\begin{itemize}
\item \textbf{Closure}: 
\[
g \bullet h \in G \quad\forall\, g, h \in G
\]
\item \textbf{Associativity}: 
\[
g \bullet (h \bullet f) = (g \bullet h) \bullet f  \quad\forall\, g,h,f \in G 
\]
\item \textbf{Identity Element}: 
\[
g \bullet e = g = e \bullet g  \quad\exists\, e\in G, \;\forall g \in G
\]
\item \textbf{Inverse Element}: 
\[
g \bullet g^{-1} = e = g^{-1} \bullet g \quad \forall g \in G,  \;\ \exists g^{-1} \in G
\]
\end{itemize}
Moreover,  the group operation $\bullet$ must satisfy the smoothness condition:
 \begin{align*}
&G \times G \rightarrow G \\
&(g, h) \mapsto g \bullet h^{-1}
 \end{align*}

\subsubsubsection{Lie Algebra}
A Lie algebra $\mathfrak{g}$ of a Lie Group $G$ is simply the tangent space $T_{e}G$ at the identity element $e$ to the Lie Group.  This vector space $\mathfrak{g}$ inherits a bilinear bracket $[-,-]$ from the group structure of $G$:
 \[
[-,-] : \mathfrak{g} \times \mathfrak{g} \rightarrow \mathfrak{g}
\]
that satisfies the following conditions for all $X, Y, Z \in \mathfrak{g}$:
\begin{itemize}
\item \textbf{Antisymmetry}:
\[
[X,  Y]  = - [Y,  X] 
\]
\item \textbf{Jacobi identity}:
\[
[X ,[Y,  Z]]  + [Y ,[Z,  X]] + [Z ,[X,  Y]] = 0
\]
\end{itemize}
These two structures – the Lie algebra and the Lie Group – allow the definition of a principal bundle and, subsequently, a connection on it. \\
A principal bundle $(P, B, G, \pi)$ is a smooth fiber bundle characterized 
by:
\begin{itemize}
  \item \textbf{Fiber:} The typical fiber is the Lie group $G$.
  \item \textbf{Right Action:} A smooth right action of $G$ on $P$:  $P \times G 
\rightarrow P$ defined by $p \cdot g$ for all $p \in P$ and $g \in G$.
\end{itemize}

This action is subject to the following constraints:
\begin{itemize}
  \item \textbf{Fiber-preserving:}
        \[
          \pi(p\cdot g)=\pi(p) \quad\forall\,p\in P,\;\forall g\in G
        \]
  \item \textbf{Free:}
        \[
          p\cdot g = p \;\Longrightarrow\; g=e \quad\forall\,p\in P,\;g\in G
        \]
  \item \textbf{Transitive on each fiber:}
        \[
          \forall\,b\in B,\;\forall\,p,q\in\pi^{-1}(b)\;
          \exists!\,g\in G \text{ such that } p\cdot g=q.
        \]
\end{itemize}

\subsubsubsection{Connection on a Principal Bundle}
A connection on a principal bundle $P = (P, B, G, \pi)$  is a connection $HP \subset TP$ on the fiber bundle $P$ characterized by the G-equivariance condition:
\[
           dR_g\!\bigl( H_pP \bigr)
           \;=\;
           H_{\,p\cdot g}P
           \qquad\forall\,p\in P,\;\forall\,g\in G,
 \]
where $R_g$ is the right translation defined as:
\[
R_g : P \rightarrow P ,\; R_g(p) := p \cdot g
\]
with its differential, $dR_g$, is the pushforward:
\[
dR_g|_p: T_pP \rightarrow T_{p \cdot g} P
\]
This can equivalently be expressed in terms of a connection 1-form $\omega$. 
\[
\omega \in \Omega^{1}(P) \otimes \mathfrak{g}
\]
where the kernel at each point is exactly the horizontal subspace:
\[
H_p P = ker(\omega_p)
\]
This $\omega$ must satisfy the following two compatibility conditions:
\begin{itemize}
\item \textbf{Reproduction on verticals}: For each $X\in\mathfrak{g}$,  the fundamental (vertical) vector field $X^\#$ (the infinitesimal generator of the G-action) obeys:
\[
\omega( X^\#) = X
\]
\item \textbf{Ad-equivariance}:
\[
R^{\ast}_g \omega = Ad(g^{-1}) \omega
\]
where $R^{\ast}_g$ is the pullback of $R_g$ and $Ad(g^{-1})$ is the adjoint action of the Lie group $G$ on the Lie algebra $\mathfrak{g}$.
\end{itemize}
The Gauge Field $A$ can be defined by selecting a local section $s$ (a gauge choice) on an open neighborhood $U$:
\[
s: U \subset B \rightarrow P
\]
where the section is defined such that its projection onto $U \subset B$ recovers the identity on $U$: 
\[
\pi_P \circ s = Id_{U} 
\]
In this case, the Gauge Field is given as the pullback of the connection 1-form $\omega$:
\[
A := s^{\ast} \omega \in \Omega^{1}(U) \otimes \mathfrak{g}
\]
Given a $G$-valued function $g: U \rightarrow G$,  the local section is transformed as:
\[
s'(x) = s(x)  g(x) \quad \forall x \in U
\]
This transformation induces a gauge transformation on the gauge field $A$ given by:
\[
A'(x) = g^{-1}(x) A(x) g(x) + g^{-1}(x)  dg(x)
\]
where:
\[
A = s^{\ast} \omega \quad A' = s'^{\ast} \omega
\]
The field strength (or curvature) $F \in \Omega^{2}(U) \otimes \mathfrak{g}$ of a gauge field $A$ is defined as:
\[
F = dA + A \wedge A
\]

\noindent For further information on differential forms and pullbacks, consult the references listed above.

\subsubsection{Associated Bundle}
An associated bundle $\hat{A}$ is constructed by “gluing” together copies of a smooth manifold $F$ using the transition data of a principal bundle $P$, in such a way that each fiber inherits a given $G$-action.  Importantly,  any connection on the principal bundle induces a natural notion of parallel transport on the associated bundle,  thereby defining a connection on $\hat{A}$ as well. This allows for a meaningful comparison of nearby fibers.\\
\noindent Mathematically,  the associated bundle $\hat{A}$ is defined as the quotient:
\[
\hat{A} = P \times_{G} F =( P \times F) / \sim
\]
with principal bundle $P = (P, B, G, \pi)$ and $F$ being a smooth manifold (or vector space) equipped with a smooth left action of Lie group $G$.\\
The associated bundle $\hat{A}$ is specified by the following:
\begin{itemize}
\item \textbf{Diagonal $G$-action}:
      The (right) $G$-action on $P\times F$ defined as:
      \[
        (p,f)\cdot g \;=\; \bigl(p\cdot g,\; g^{-1}\!\cdot\! f\bigr)
      \]
      This action is free and proper.
\item \textbf{Equivalence relation}:
	The equivalence relation $\sim$ is defined as:
	\[
		 (p,\;f)\sim(p\!\cdot\!g,\;g^{-1}\!\cdot\!f)\qquad
		 \forall\,p\in P,\;g\in G,\;f\in F .
	\]
\item \textbf{Quotient Map}:
	The quotient map $q$ is specified by:
	\[
		q: P \times F \rightarrow \hat A , \; (p, f) \mapsto [p, f]
	\]	
\item \textbf{Local trivialisations}:
	For every open set $U\subset B$ with a local section
        $s:U\to P$,  the map
        \[
            \Phi_{s}:U\times F\longrightarrow\pi_{\hat A}^{-1}(U),\quad
            (x,f)\longmapsto[s(x),f]
        \]
        is a diffeomorphism.  The associated bundle projection $\pi_{\hat A}$ is defined as:
	\[
           \pi_{\hat A}: \hat A \rightarrow B, \; \pi_{\hat{A} }([p, f]) = \pi(p)
        \]
	Consequently, $\hat A$ is locally isomorphic to the
        product bundle $U\times F$.  When $F$ is a vector space and the
        $G$-action is linear, these trivialisations endow $\hat A$ with the
        natural structure of a vector bundle.
\end{itemize}
The construction of an associated bundle can be illustrated by the following commutative diagram:
\[
\begin{tikzcd}[row sep=large, column sep=huge]
P \times F \arrow[r,"q"] \arrow[d,"\mathrm{pr}_P"'] 
  & \hat{A} = P\times_G F \arrow[d,"\pi_{\hat{A} }"] \\
P \arrow[r,"\pi"'] 
  & B
\end{tikzcd}
\]
where $pr_P: P \times F \rightarrow P$ represent the projection onto the first factor.

\section{Gauge Flow Model}
A Gauge Flow Model is defined within the framework of the associated bundle $\hat{A}$:
\[
 	\hat{A} = P \times_{G} F
\]
with the following:
\begin{itemize}
\item The principal bundle $P = (P, M, G, \pi_P)$ with Lie group $G$ and base manifold $M$.
\item The typical fiber space $F$ with a smooth left action of the Lie group $G$.
\end{itemize}
\noindent The dynamics of a Gauge Flow Model are governed by the subsequent differential equation:
\[
\hat{\nabla}_{dt} x(t):= v_{\theta}(x(t),  t) - \alpha(t) \Pi_{M} ( A_{\mu}(x(t),  t) d^{\mu}(x(t), t)  v_{s}(x(t), t) )
\]
Here,  the components are defined as follows:
\begin{itemize}
\item $v_{\theta}(x(t),  t) \in TM$ is a learnable vector field modeled by a Neural Network.
\item $\alpha (t)$ is a time dependent weight,  also modeled by a Neural Network.
\item $A$ is the learnable gauge field corresponding to the connection on the principal bundle $P$, also represented by a Neural Network.
\item $d^{\mu}(x(t), t) \in TM$ is the direction vector field.
\item $v_{s}(x(t), t) : M \rightarrow P \times_G F$  is the fiber section,  which can be learned by a Neural Network.
\item $\Pi_{M}: P \times_G F \rightarrow TM$ is a smooth projection map from the associated bundle $\hat{A} = P \times_G F$ to the tangent bundle $TM$. This map covers the identity on $M$, ensuring that any element $[p, v] \in P \times_G F$ is projected to the tangent space of its corresponding base point in $M$:
\[
\pi_{TM} ( \Pi_{M}([p,v])) = \pi_{\hat{A}}([p,v]) \quad \forall [p,v] \in P \times_G F
\]
\end{itemize}
\noindent The Gauge Flow Model is defined for any differentiable base manifold $M$.
However,  training the model requires $M$ to admit the structure of 
a Riemannian manifold, specifically possessing a Riemannian metric $g$
defined on it.\\
\noindent Training a Gauge Flow Model employs the Riemannian Flow Matching (RFM) 
framework~\cite{RFM},  which generalizes ordinary Flow Matching 
(FM)~\cite{FM}. The training objective is the Gauge Flow Model loss, 
defined as:
\[
\mathcal{L}_{\mathrm{GFM}} =  \mathbb{E}_{\substack{t \sim \mathcal{U}[0,1] \\ x \sim p_t}}  \bigg\lVert \Big[ v_{\theta}(x,  t) - \alpha(t) \Pi_{M} ( A_{\mu}(x,  t) d^{\mu}(x, t)  v_{s}(x, t) )\Big] - u_{t}(x) \bigg\rVert_{g_x}^{2}
\]
where $\|\cdot\|_{g_x}$ is the norm induced by the Riemannian metric $g$ of the base manifold $M$ at point $x$.
Here,  both terms lie in the tangent space of the base manifold $M$:
\begin{gather*}
\Big[ v_{\theta}(x,  t) - \alpha(t) \Pi_{M} ( A_{\mu}(x,  t) d^{\mu}(x, t)  v_{s}(x, t) )\Big] \in T_x M\\
 u_{t}(x) \in T_x M 
\end{gather*}
\newpage
The Gauge Flow Model loss incorporates the following components:
\begin{itemize}
  \item \textbf{Probability-density path}  
      $p_t : M \to \mathbb R_{+}$, $t \in [0,1]$, as defined in~\cite{RFM}.  
      For each~$t$, $p_t$ satisfies
      \[
        \int_{M} p_t(x)\, d\mathrm{vol}_x = 1,
      \]
      where $d\mathrm{vol}_x$ is the Riemannian volume form.
  \item \textbf{Target vector field}  
      $u_t(x)$, obtained from the conditional field $u_t(x \mid x_1)$ by
      \[
        u_t(x) =
        \int_{M}
        u_t\!\bigl(x \mid x_1\bigr)\,
        \frac{p_t\!\bigl(x \mid x_1\bigr)\, q(x_1)}{p_t(x)}
        \, d\mathrm{vol}_{x_1},
      \]
      with $q(x) \coloneqq p_{t=1}(x)$ and $x_1$ denoting the sample at time $t=1$.
\end{itemize}
\noindent Computing the Gauge Flow Model (GFM) loss $\mathcal{L}_{\mathrm{GFM}}$ exactly is generally intractable. However, this loss is equivalent to the Riemannian Flow Matching (RFM) objective. To address this intractability,~\cite{RFM} introduces the Riemannian Conditional Flow Matching (RCFM) loss—a single-sample, unbiased Monte Carlo estimator of the RFM objective. The key idea is to replace the marginal target field $u_t(x)$ with its conditional counterpart $u_t(x \mid x_1)$, enabling tractable training. This formulation allows for efficient, simulation-free learning on simple geometries and scalable neural network training on general Riemannian manifolds.  \hspace*{-0.2cm}Notably,  this approach can also be applied to train Gauge Flow Models.

\noindent For full derivations and implementation details,  refer to~\cite{RFM,FM}.

\section{Experiments}
This study compares the novel Gauge Flow Models to standard Flow Models 
using a generated Gaussian Mixture Model (GMM) dataset.  The Gauge Flow Models were evaluated for several values of $N$ associated with the Lie Group $G = SO(N)$.  
\subsection{Model}
The configuration of the Gauge Flow Models consists of the following components:
\begin{itemize}
\item The trivial principal bundle $P = (P= M\times G,  M,  G,\pi)$ with structure Lie group $G=SO(N)$ and base manifold $M = \mathbb{R}^{N}$.
\item The typical fiber $F = \mathbb{R}^{N}$.  
\item The vector field $v_{\theta} (x(t), t)$ defined on the tangent bundle $TM$,  is modeled by a Neural Network.
\item The alpha function $\alpha(t)$ is also modeled using a Neural Network.
\item The gauge field $A \in \Omega^{1} (M) \otimes \mathfrak {g}$, valued in the Lie algebra $\mathfrak{g} = \mathfrak{so}(N)$ of the Lie group $G= SO(N)$,  is represented by a Neural Network.
\item The directional vector field $d^{\mu} (x(t), t)$ is either identified with the vector field $v_{\theta} (x(t), t)$ or $v_{s}(x(t), t)$ leading to two distinct Gauge Flow Model variants.
\item The vector field $v_{s}(x(t), t) : M \rightarrow P \times_G F$ is expressed in the gauge defined by orthonormal bases of $\mathbb{R}^N$.  In this gauge,  the field $v_{s}(x(t), t)$ corresponds to a global map $M \rightarrow F$,  which is modeled using a Neural Network.
\item The mapping $ \Pi_{M} : P \times_G \mathbb{R}^{N} \rightarrow M \times \mathbb{R}^{N}$ acts as the identity function on both $M$ and $\mathbb{R}^N$,  with the $G$-component effectively projected out as a consequence of the chosen gauge.
\end{itemize}
The Neural Network models used are standard Multi-Layer Perceptrons (MLPs) with the subsequent configurations:
\begin{table}[h!]
\centering
\begin{tabular}{|c|c|c|}
  \hline
  Field & Layer Dimensions & Activation Function \\
  \hline
   $v_{\theta} (x(t), t)$      &   [N + 1,   32/64,  32/64,  N]    &  SiLU        \\
 $A$       &   [N + 1,   32/64,  32/64,  $N^2 (N-1) //2$ ]    &  SiLU         \\
  $v_{s}(x(t), t)$       &   [N + 1,   32/64,  32/64,  N]    &  SiLU         \\
 $\alpha(t)$       &   [1,   16,   1 ]    &  SiLU         \\
  \hline
\end{tabular}
\end{table}\\
The layer width depends on the dimension $N$ and is set to 32 for $N > 10$ and to 64 otherwise.  Here,  SiLU refers to the Sigmoid Linear Unit activation function. The Plain Flow Model is implemented as a multilayer perceptron (MLP),  where the number of layers and their widths are functions of the input dimension $N$.  Initially,  the network architecture follows the layer dimensions $[N + 1, 128, 128, 128, N]$,  with SiLU used as the activation function throughout. These Gauge Flow Models are generally trained within the framework of Riemannian Flow Matching \cite{RFM}.  However,  since the base manifold is $\mathbb{R}^{N}$, training can be performed using the standard Flow Matching \cite{FM}.
\subsection{Dataset}
The datasets used in the experiment are generated Gaussian-Mixture Model (GMM) datasets with:
\[
  n_{\text{train}} = 15{,}000,\qquad
  n_{\text{test}}  = 5{,}000
\]
i.i.d.\ samples from a mixture of \(K=3000\) equally-weighted Gaussian components in ambient dimension \(N\in\{3,\dots,32\}\).
The GMM datasets have the following specifications:
\begin{itemize}
  \item \textbf{Mixture weights:} \(\pi_k = \frac1K,\; k=0,\dots,K-1\).
  \item \textbf{Covariances:} \(\Sigma_k = 0.5\,I_N\) (isotropic).
  \item \textbf{Spread parameter:} \(\alpha = 25\).
  \item \textbf{Component means:} $\mu_k$ for each component index \(k\)
        \begin{enumerate}
          \item \emph{Primary axis:}
                \(a_1 = k \bmod N,\;
                  \mu_{k,a_1} = (-1)^k \,\alpha.\)
          \item \emph{Secondary axis:}
                \(a_2 = (k + \lfloor K/2\rfloor) \bmod N.\)
                If \(a_2\neq a_1\) set
                \(\mu_{k,a_2} = (-1)^{k+1}\,\tfrac12\alpha.\)
          \item \emph{Extra offset} (only when \(K>N\) and \(k\ge N\)): \\
                let \(b=(a_1+\lfloor k/N\rfloor)\bmod N\) and add
                \[
                  \mu_{k,b} \;{+}= \;
                  s_k\,0.1\,\alpha \,\lfloor k/N\rfloor,\qquad
                  s_k=\begin{cases}
                    +1,& k\bmod3=0,\\
                    -1,& \text{otherwise}.
                  \end{cases}
                \]
        \end{enumerate}
\end{itemize}

\noindent
\textbf{Sampling:}\; Draw \(k\sim\mathrm{Cat}(\pi)\) and
\(x\sim\mathcal N\!\bigl(\mu_k,\,0.5\,I_N\bigr).\)

\subsection{Results}
The performance of the models on the GMM dataset was evaluated using both 
training and testing loss,  along with the number of parameters. 

\subsubsection{Train Loss}
The Gauge Flow Model utilizing the directional vector field $v_{\theta} 
(x(t), t)$ or $v_{s}(x(t), t)$,  demonstrated superior training 
performance.  To facilitate meaningful comparisons,  the training loss was 
normalized relative to the loss of the $v_{\theta} (x(t), t)$ Gauge Flow 
Models.  Notably,  the training performance varied across different Lie 
group dimensions ($N$s),  with the Gauge Flow Model using $v_{\theta} (x(t), 
t)$ consistently exhibiting the best results for most values of $N$.
\begin{figure}[h!]
  \centering
  \includegraphics[width=0.8\textwidth]{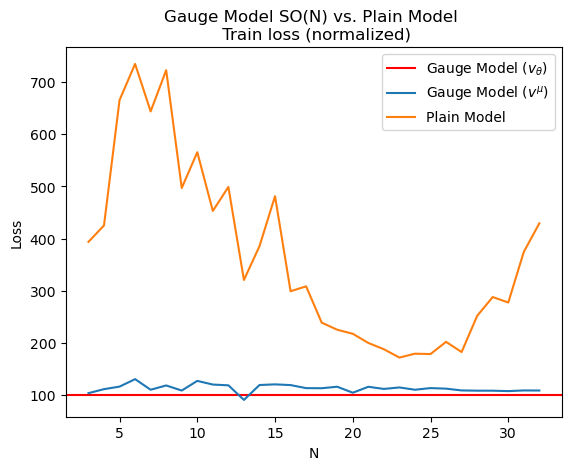}
  \caption{Train Loss Comparison (lower is better): The normalized train loss is plotted against 
the dimension $N$ to compare the performance of the Gauge Flow Model 
using directional vector fields $v_{\theta}(x(t), t)$ and $v_{s}(x(t), 
t)$ with the standard plain Flow Model.}
  \label{fig:TrainLoss}
\end{figure}
\subsubsection{Test Loss}
The test loss exhibited similar performance trends to the train loss.  
Both Gauge Flow Model variants significantly outperformed the standard Flow Model across all values of $N$.
\begin{figure}[h!]
  \centering
  \includegraphics[width=0.8\textwidth]{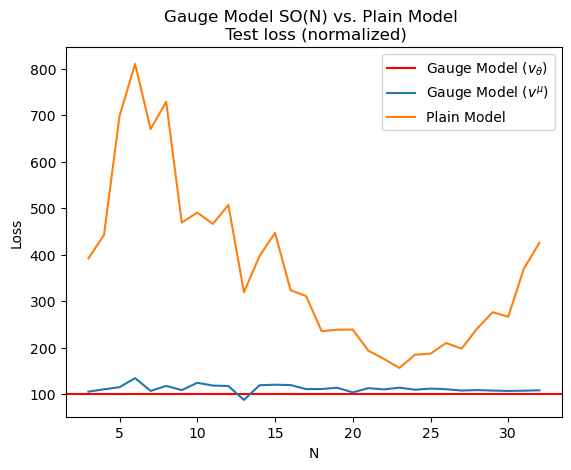}
  \caption{Test Loss Comparison (lower is better): The normalized test loss is plotted against 
the dimension $N$ to compare the performance of the Gauge Flow Model 
using directional vector fields $v_{\theta}(x(t), t)$ and $v_{s}(x(t), 
t)$ with the standard plain Flow Model. }
  \label{fig:TestLoss}
\end{figure}
\\
\subsubsection{Number of Parameters}
\begin{figure}[h!]
  \centering
  \includegraphics[width=0.8\textwidth]{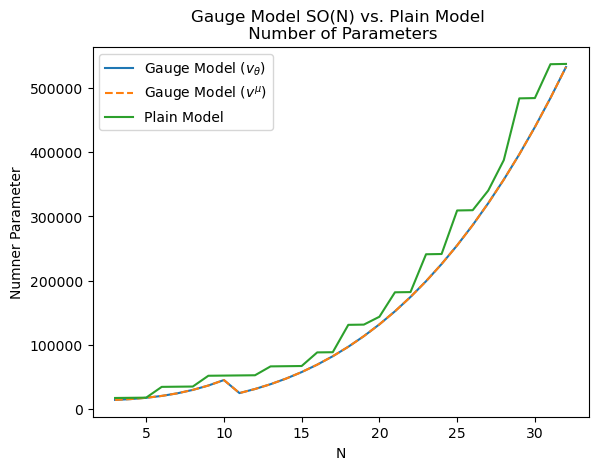}
  \caption{Number of Parameters: A comparison of the number of parameters for the 
plain Flow Model and the Gauge Flow Models across various dimensions $N$.}
  \label{fig:NumParams}
\end{figure}

As shown in Figure~\ref{fig:NumParams}, the plain Flow Model had slightly more parameters than the Gauge Flow Models across all dimensions $N$,  but the two variants ultimately had the same number of parameters.

\newpage
\section{Related Work}
The concept of Continuous Normalizing Flows (CNFs),  as introduced in \cite{CNFs},  traditionally relies on computationally intensive Ordinary Differential Equation (ODE) solvers during training,  guided by a maximum likelihood objective.  While effective,  this training procedure scales poorly due to the repeated integration of ODEs.  To mitigate this limitation, the Flow Matching paradigm was proposed in \cite{FM}.  Instead of maximizing likelihood,  Flow Matching formulates training as an $L_2$ regression problem that directly aligns a learnable vector field with a target velocity field.   This target velocity is derived from known boundary conditions,  significantly reducing computational overhead during training.  Initially formulated in Euclidean space,  Flow Matching was later extended to Riemannian manifolds in the Riemannian Flow Matching (RFM) framework \cite{RFM}.  In RFM,  the vector field matching operates on curved spaces while respecting the geometry induced by a fixed Riemannian metric.  However,  it is important to note that the connections—i.e., the geometric structures defining parallel transport and differentiation—are determined by the manifold and are not themselves learnable parameters.\\
\noindent Parallel to these developments,  several lines of research have explored the connections between gauge theory and deep learning.  A notable example is presented in \cite{GNN},  where a gauge-invariant framework is introduced that encodes invariance directly into the inputs by constructing gauge-invariant features before passing them through standard neural architectures.  This approach imposes gauge symmetry directly on the inputs instead of embedding it in the model dynamics.\\
Another approach leveraging geometric and gauge-theoretic principles is presented in \cite{GCNNs},  where standard Convolutional Neural Networks (CNNs) are generalized to operate on arbitrary manifolds by enforcing coordinate independence.  This work builds upon the foundations laid by steerable CNNs \cite{SteerableCNNs} and Group Equivariant CNNs (G-CNNs) \cite{GECNNs},  extending the principle of equivariance to manifold-valued data.
Further advancing this direction,  Lattice Gauge Equivariant Convolutional Neural Networks (L-CNNs) \cite{LGCNN} incorporate local gauge symmetry directly into the network architecture.  These models integrate gauge equivariance into both convolutional layers and bilinear operations used to compute Wilson loops,  which are gauge-invariant observables from lattice gauge theory.  This framework was subsequently extended in \cite{GLCNN} to include global lattice symmetries,  such as rotations and reflections,  thereby achieving a comprehensive symmetry-aware neural network design.

\printbibliography

\end{document}